# Old Content and Modern Tools – Searching Named Entities in a Finnish OCRed Historical Newspaper Collection 1771–1910

Kimmo Kettunen[1], Eetu Mäkelä[2], Teemu Ruokolainen[3], Juha Kuokkala[4] and Laura Löfberg[5]

[1] National Library of Finland, Centre for Preservation and Digitization, Mikkeli, Finland
kimmo.kettunen@helsinki.fi
[2] Aalto University, Semantic Computing Research Group, Espoo, Finland
eetu.makela@aalto.fi
[3] National Library of Finland, Centre for Preservation and Digitization, Mikkeli, Finland
teemu.ruokolainen@helsinki.fi
[4] University of Helsinki, Department of Modern Languages, Helsinki, Finland
juha.kuokkala@helsinki.fi
[5] Department of Linguistics and English Language,
Lancaster University, UK
l.lofberg@lancaster.ac.uk

Named Entity Recognition (NER), search, classification and tagging of names and name like frequent informational elements in texts, has become a standard information extraction procedure for textual data. NER has been applied to many types of texts and different types of entities: newspapers, fiction, historical records, persons, locations, chemical compounds, protein families, animals etc. In general a NER system's performance is genre and domain dependent and also used entity categories vary (Nadeau and Sekine, 2007). The most general set of named entities is usually some version of three partite categorization of locations, persons and organizations. In this paper we report first large scale trials and evaluation of NER with data out of a digitized Finnish historical newspaper collection Digi. Experiments, results and discussion of this research serve development of the Web collection of historical Finnish newspapers.

Digi collection contains 1,960,921 pages of newspaper material from years 1771–1910 both in Finnish and Swedish. We use only material of Finnish documents in our evaluation. The OCRed newspaper collection has lots of OCR errors; its estimated word level correctness is about 70–75 % (Kettunen and Pääkkönen, 2016). Our principal NER tagger is a rule-based tagger of Finnish, FiNER, provided by the FIN-CLARIN consortium. We show also results of limited category semantic tagging with tools of the Semantic Computing Research Group (SeCo) of the Aalto University. Three other tools are also evaluated briefly.

This research reports first published large scale results of NER in a historical Finnish OCRed newspaper collection. Results of the research supplement NER results of other languages with similar noisy data. As the results are also achieved with a small and morphologically rich language, they illuminate relatively well researched area of Named Entity Recognition from a new perspective.

**Keywords:** named entity recognition, historical newspaper collections, Finnish

# 1    Introduction

The National Library of Finland has digitized a large proportion of the historical newspapers published in Finland between 1771 and 1910 (Bremer-Laamanen, 2014; Kettunen *et al.*, 2014). This collection contains 1,960,921 million pages in Finnish and Swedish. Finnish part of the collection consists of about 2.4 billion words. The National Library's Digital Collections are offered via the *digi.kansalliskirjasto.fi* web service, also known as Digi. Part of the newspaper material (years 1771–1874) is freely downloadable in The Language Bank of Finland provided by the FIN-CLARIN consortium[1]. The collection can also be accessed through the Korp[2] environment that has been developed by Språkbanken at the University of Gothenburg and extended by FIN-CLARIN team at the University of Helsinki to provide concordances of text resources. A Cranfield style information retrieval test collection has been produced out of a small part of the Digi newspaper material at the University of Tampere (Järvelin *et al.*, 2015). An open data package of the whole collection will be released during the year 2016 (Pääkkönen *et al.*, 2016).

The web service digi.kansalliskirjasto.fi is used, for example, by genealogists, heritage societies, researchers, and history enthusiast laymen. There is also an increasing desire to offer the material more widely for educational use. In 2015 the service had about 14 million page loads. User statistics of 2014 showed that about 88.5 % of the usage of the Digi came from Finland, but an 11.5 % share of use was coming outside of Finland.

Digi is part of the growing global network of digitized newspapers and journals, and historical newspapers are considered more and more as an important source of historical knowledge. As the amount of digitized data grows, also tools for harvesting the data are needed to gather information. Named Entity Recognition has become one of the basic techniques for information extraction of texts since mid-1990's (Nadeau and Sekine, 2007). In its initial form NER was used to find and mark semantic entities like person, location and organization in texts to enable information extraction related to these kinds of entities. Later on other types of extractable entities, like time, artefact, event and measure/numerical, have been added to the repertoires of NER software (Nadeau and Sekine, 2007; Kokkinakis *et al.*, 2014).

Our goal with usage of NER is to provide users of Digi better means for searching and browsing the historical newspapers, i.e. new ways to structure, access and possi-bly also enhance information. Different types of names, especially person names and names of locations are used frequently as search terms in different newspaper collections (Crane and Jones, 2006). They can provide also browsing assistance to collections, if the names are recognized and tagged in the newspaper data and put into the index (Neudecker *et al.*, 2014). A fine example of usage of name recognition with historical newspapers is La Stampa's historical newspaper collection[3]. After basic keyword search users can browse or filter the search results by using three basic NER categories of person (authors of articles or persons mentioned in the articles), location (countries and cities mentioned in the articles) and organization. Thus named entity annotations of newspaper text allow a more semantically-oriented exploration of content of the large archive. Another large scale (152 M articles) NER analysis of the Australian historical newspaper collection Trove with usage examples is described in Mac Kim and Cassidy (2015).

---

[1] https://kitwiki.csc.fi/twiki/bin/view/FinCLARIN/KielipankkiAineistotDigilibPub
[2] https://korp.csc.fi/

Our main research question in this article is, how well or poorly names can be recognized in an OCRed historical Finnish newspaper collection with available software tools. The task has many pitfalls that will affect the results: firstly, the word level quality of the material is quite low (Kettunen and Pääkkönen, 2016). Secondly, we have available only language technology tools that are made for modern Finnish. Thirdly, there is no available comparable NER data of Finnish, neither a standard evaluation corpus. Thus our results form a first baseline for NER of historical Finnish. It is expectable that results will not be very good, but they will give us a realistic empirical perspective on NER's usability with our data.

We shall not provide a review of basic NER literature; those who are interested in getting an overall picture of the topic, can start e.g. with Nadeau and Sekine (2007), who offer both historical and methodological basics of the theme. References in Nadeau and Sekine and in this paper allow further familiarization. Specific problems related to historical language and OCR problems are discussed in the paper in relation to our data.

The structure of the paper is following: first we introduce our NER tools, our evaluation data and the tag set. Then we'll show results of evaluations and finally discuss the results and our plans for usage of NER with the on-line newspaper collection.

## 2 NER Software and Evaluation

For recognition and labelling of named entities in our evaluation we use principally FiNER software. SeCo's ARPA is a different type of tool, it is mainly used for Semantic Web tagging and linking of entities (Mäkelä, 2014)[4], but it could be adapted for basic NER, too. Besides these two tools, three others were also evaluated briefly. Connexor[5] has NER for modern Finnish, which is commercial software. Multilingual package Polyglot[6] works also for Finnish and recognizes persons, places and organi-

---

zations. A semantic tagger for Finnish (Löfberg *et al.*, 2005) recognizes also three types of names.

Both FiNER and ARPA have been implemented as analysers of modern Finnish, although ARPA's morphological engine is able to deal with phenomena of 19[th] century Finnish, too. As far as we know there is so far no NER tagger for historical Finnish available. Before choosing FiNER and ARPA we tried also a commonly used trainable free tagger, Stanford NER[7], but were not able to get reasonable performance out of it for our purposes, although the software has been used successfully for other languages than English, too. Dutch, French and German named entity recognition with the Stanford NER tool has been reported in the Europeana historical newspaper project, and the results have been good (Neudecker *et al.*, 2014; Neudecker, 2016).

## 2.1 FiNER

FiNER is a rule-based named-entity tagger, which in addition to surface text forms utilizes grammatical and lexical information from a morphological analyzer (Omorfi[8]). FiNER pre-processes the input text with a morphological tagger derived from Omorfi. The tagger disambiguates Omorfi's output by selecting the statistically most probable morphological analysis for each word token, and for tokens not recognized by the analyzer, guesses an analysis by analogy of word-forms with similar ending in the morphological dictionary. The use of morphological pre-processing is crucial in performing NER with a morphologically rich language such as Finnish (and Estonian (Tkachenko *et al.*, 2013)), where a single lexeme may theoretically have thousands of different inflectional forms.

The focus of FiNER is in recognizing different types of proper names. Additionally, it can identify the majority of Finnish expressions of time and e.g. sums of money. FiNER uses multiple strategies in its recognition task:

1) Pre-defined gazetteer information of known names of certain types. This information is mainly stored in the morphological lexicon as additional data tags of the lexemes in question. In the case of names consisting of multiple words, FiNER rules incorporate a list of known names not caught by the more general rules.

2) Several kinds of pattern rules are being used to recognize both single- and multiple-word names based on their internal structure. This typically involves (strings of) capitalized words ending with a characteristic suffix such as Inc, Corp, Institute etc. Morphological information is also utilized in avoiding erroneously long matches, since in most cases only the last part of a multi-word name is inflected, while the other words remain in the nominative (or genitive) case. Thus, preceding capitalized words in other case forms should be left out of a multi-word name match.

3) Context rules are based on lexical collocations, i.e. certain words which typically or exclusively appear next to certain types of names in text. For example, a string of capitalized words can be inferred to be a corporation/organization if it is followed by a verb such as *tuottaa* ('produce'), *työllistää* ('employ') or *lanseerata* ('launch' [a product]), or a personal name if it is followed by a comma- or parenthesis-separated numerical age or an abbreviation for a political party member.

The pattern-matching engine that FiNER uses, HFST Pmatch, marks leftmost longest non-overlapping matches satisfying the rule set (basically a large set of disjuncted patterns) (Linden *et al.*, 2013; Silfverberg, 2015). In the case of two or more rules matching the exact same passage in the text, the choice of the matching rule is undefined. Therefore, more control is needed in some cases. Since HFST Pmatch did not contain a rule weighing mechanism at the time of designing the first release of FiNER, the problem was solved by applying two runs of distinct Pmatch rulesets in succession. This solves for instance the frequent case of Finnish place names used as family names: in the first phase, words tagged lexically as place names but matching a personal name context pattern are tagged as personal names, and the remaining place name candidates are tagged as places in the second phase. FiNER annotates 15 different entities that belong to five semantic categories: location, person, organization, measure and time (Silfverberg, 2015).

## 2.2 ARPA

SeCo's ARPA (Mäkelä, 2014) is not actually a NER tool, but instead a dynamic, configurable entity linker. In effect, ARPA is not interested in locating all entities of a particular type in a text, but instead locating all entities that can be linked to strong identifiers elsewhere. Through these, it is then for example possible to source coordinates for identified places, or associate different name variants and spellings to a single individual. For the pure entity recognition task presented in this paper, ARPA is thus at a disadvantage. However, we wanted to see how it would fare in comparison to FiNER.

The core benefits of the ARPA system lie in its dynamic, configurable nature. In processing, ARPA combines a separate lexical processing step with a configurable SPARQL-query -based lookup against an entity lexicon stored at a Linked Data endpoint. Lexical processing for Finnish is done with a modified version of Omorfi[9], which supports historical morphological variants, as well as lemma guessing for out of vocabulary words. This separation of concerns allows the system to be speedily configured for both new reference vocabularies as well as the particular dataset to be processed.

## 2.3 Evaluation Data

As there was no evaluation collection for Named Entity Recognition of 19[th] century Finnish, we needed first to create one. As evaluation data we used samples from different decades out of the Digi collection. Kettunen and Pääkkönen (2016) calculated among other things number of words in the data for different decades. It turned out that most of the newspaper data was published in 1870–1910, and beginning and mid of the 19[th] century had much less published material. About 95 % of the material was printed in 1870–1910, and most of it, 82.7 %, in the two decades of 1890–1910.

---

[9] https://github.com/jiemakel/omorfi

We aimed at an evaluation collection of 150,000 words. To emphasize the importance of the 1870–1910 material we took 50 K of data from time period 1900–1910, 10 K from 1890–1899, 10 K from 1880–1889, and 10 K from 1870–1879. Rest 70 K of the material was picked from time period of 1820–1869. Thus the collection reflects most of the data from the century but is also weighed to the end of the 19[th] century and beginning of 20[th] century. Decade-by-decade word recognition rates in Kettunen and Pääkkönen (2016) show that word recognition rate during the whole 19[th] century is quite even. Thus we believe that temporal dimension of the data should not bring great variation to the NER results. It may be possible, however, that older data has old names that are out of FiNER's scope.

The final manually tagged evaluation data consists of 75,931 lines, each line having one word or other character data. By character data we mean here that line contains misrecognized words that have a variable amount of OCR errors. The word accuracy of the evaluation sample is on the same level as the whole newspaper collection's word level quality: about 73 % of the words in the evaluation collection can be recognized by a modern Finnish morphological analyzer. The recognition rate in the whole index of the newspaper collection is estimated to be in the range of 70–75 % (Kettunen and Pääkkönen, 2016). Evaluation data was input to FiNER as small textual snippets. 71 % of the tagger's input snippets have five or more words, the rest have fewer than five words in the text snippet. Thus the amount of context the tagger can use in recognition is varying.

FiNER uses 15 tags for different types of entities, which is too fine a distinction for our purposes. Our first aim was to concentrate only on locations and person names, because they are mostly used in searches of the Digi collection, as was detected in an earlier log analysis, where 80 % of the ca. 149 000 occurrences of top 1000 search term types consisted of first and last names of persons and place names (Kettunen *et al.*, 2014). This kind of search term use is very common especially in the humanities information seeking (Crane and Jones, 2006).

After reviewing some of the FiNER tagged material, we included also three other tags, as they seemed important and were occurring frequently enough in the material. The eight final chosen tags are shown and explained below.

| Entity/tag | Meaning |
| --- | --- |
| 1. <EnamexPrsHum> | person |
| 2. <EnamexLocXxx> | general location |
| 3. <EnamexLocGpl> | geographical location |
| 4. <EnamexLocPpl> | political location (state, city etc.) |
| 5. <EnamexLocStr> | street, road, street address |
| 6. <EnamexOrgEdu> | educational organization |
| 7. <EnamexOrgCrp> | company, society, union etc. |
| 8. <TimexTmeDat> | expression of time |

The final entities show that our interest is mainly in the three most generally used semantic NER categories: persons, locations and organizations. In locations we have four different categories and with organizations two. Temporal expressions were included in the tag set due to their general interest in the newspaper material.

Manual tagging of the evaluation corpus was done by the third author, who had previous experience in tagging modern Finnish with tags of the FiNER tagger. Tagging took one month, and quality of the tagging and its principles were discussed before starting based on a sample of 2000 lines of evaluation data. It was agreed, for example, that words that are misspelled but are recognizable for the human tagger as named entities would be tagged (cf. 50 % character correctness rule in Packer *et al.*, 2010). If orthography of the word was following 19th century spelling rules, but the word was identifiable as a named entity, it would be tagged, too.

To get an idea how well FiNER recognizes names in general, we evaluated it with a list of 75 980 names of locations and persons. We included in the lists modern first names and surnames, old first names from the 19th century, names of municipalities, and names of villages and houses. The list contains also names in Swedish, as Swedish was the dominant language in Finland during most of the 19[th] century. The list has been compiled from independent sources that include e.g. Institute for the Languages of Finland, National Land Survey of Finland, Genealogical Society of Finland, among others. All the names were given to FiNER as part of a predicative pseudo sentence *X on mukava juttu* ('X is a nice thing') so that the tagger had some context to work with, not just a list of names.

FiNER recognized 55,430 names out of the list, 72.96 %. Out of these 8,904 were tagged as persons, 35,733 as LocXxxs, and 10,408 as LocGpls. The rest were tagged as organizations, streets, time and title. Among locations FiNER favors general locations, LocXxxs. As LocGpls it tags locations that have some clear mention of a natural geographical entity as part of the name (lake, pond, river, hill, rapid etc.), but this is not clear cut, as some names of this type seem to get tag of LocXxx. It would be reasonable to use only one location tag with FiNER, as the differences between location categories are not very significant.

Among the names that FiNER does not recognize are foreign names, mostly Swedish (also in Sami), names that can also be common nouns, different compound names, and old names. Variation of *w/v*, one the most salient differences of 19[th] century Finnish and modern Finnish, does not impair FiNER's tagging, although it has a clear impact on general recognizability of 19[th] century Finnish (Kettunen and Pääkkönen, 2016). Some other differing morphological features of 19[th] century Finnish (Järvelin *et al.*, 2015, cf. Table 1) may affect recognition of names with FiNER.

## 2.4    Results of the Evaluation

We evaluated performance of FiNER and SeCo's ARPA using the *conlleval*[10] script used in Conference on Computational Natural Language Learning (CONLL).

---

[10]   http://www.cnts.ua.ac.be/conll2002/ner/bin/conlleval.txt, author ErikTjong Kim Sang, version 2004-01-26

*Conlleval* uses standard measures of precision, recall and F-score, the last one defined as 2PR/(R+P), where P is precision and R recall (cf. Manning and Schütze, 1999: 269). Evaluation is based on "exact-match evaluation" (Nadeau and Sekine, 2007). In this type of evaluation NER system is evaluated based on the micro-averaged F-measure (MAF) where precision is the percentage of correct named entities found by the NER software; recall is the percentage of correct named entities present in the tagged evaluation corpus that are found by the NER system. A named entity is considered correct only if it is an exact match of the corresponding entity in the tagged evaluation corpus: "a result is considered correct only if the boundaries and classification are exactly as annotated" (Poibeau and Kosseim, 2001). Thus the evaluation criteria are strict, especially for multipart entities.

## 2.5    Results of FiNER

Detailed results of the evaluation of FiNER are shown in Table 1. Entities <ent/> consist of one word token, <ent> are part of a multiword entity and </ent> are last parts of multiword entities.

| Label | P | R | F-score | Number of tags found | Number of tags in the evaluation data |
|---|---|---|---|---|---|
| <EnamexLocGpl/> | 6.96 | 9.41 | 8.00 | 115 | 85 |
| <EnamexLocPpl/> | 89.50 | 8.46 | 15.46 | 181 | 1920 |
| <EnamexLocStr/> | 23.33 | 50.00 | 31.82 | 30 | 14 |
| <EnamexLocStr> | 100.00 | 13.83 | 24.30 | 13 | 94 |
| </EnamexLocStr> | 100.00 | 18.31 | 30.95 | 13 | 71 |
| <EnamexOrgCrp/> | 2.39 | 6.62 | 3.52 | 376 | 155 |
| <EnamexOrgCrp> | 44.74 | 25.99 | 32.88 | 190 | 338 |
| </EnamexOrgCrp> | 40.74 | 31.95 | 35.81 | 189 | 250 |
| <EnamexOrgEdu> | 48.28 | 40.00 | 43.75 | 29 | 35 |
| </EnamexOrgEdu> | 55.17 | 64.00 | 59.26 | 29 | 25 |
| <EnamexPrsHum/> | 16.38 | 52.93 | 25.02 | 1819 | 564 |
| <EnamexPrsHum> | 87.44 | 26.67 | 40.88 | 438 | 1436 |
| </EnamexPrsHum> | 82.88 | 31.62 | 45.78 | 438 | 1150 |
| <TimexTmeDat/> | 5.45 | 14.75 | 7.96 | 495 | 183 |
| <TimexTmeDat> | 68.54 | 2.14 | 4.14 | 89 | 2857 |
| </TimexTmeDat> | 20.22 | 2.00 | 3.65 | 89 | 898 |

**Table 1.** Evaluation results of FiNER with strict CONLL evaluation criteria. Data with zero P/R is not included in the table. These include categories <EnamexLocGpl>, </EnamexLocGpl>, <EnamexLocPpl>, </EnamexLocPpl>, <EnamexLocXxx>, <Enamex-LocXxx/>, </EnamexLocXxx>, and <EnamexOrgEdu/>. Most of these have very few entities in the data, only <EnamexLocXxx> is frequent with over 1200 occurences

Results of the evaluation show that named entities are not recognized very well by FiNER, which is not surprising, as the quality of the text data is quite low. Especially recognition of multipart entities is mostly very low. Some part of the entities may be recognized, but rest is not. Out of multiword entities person names and educational organizations are recognized best. Names of persons are the most frequent category. Recall of one part person names is best, but its precision is low. Multipart person names have a more balanced recall and precision, and their F-score is 40–45. If the three different locations (*LocGpl, LocPpl* and *LocXxx*) are joined in strict evaluation as one general location, *LocXxx*, one part locations get precision of 65.69, recall of 50.27 and F-score of 56.96 with 1533 tags. Multipart locations are found badly even then. FiNER seems to have a tendency to tag most of the *LocPpl*s as *LocXxx*s. *LocG-pl*s are also favored instead of *LocPpl*s. On the other hand, only one general location like LocXxx could be enough for our purposes, and these results are reasonably good

Closer examination of street results shows that problems in street name recognition are due to three main reasons: OCR errors in street names, abbreviated street names and multipart street names with numbers as part of the name. In principle streets are easy to recognize in Finnish, while they have most of the time common part *katu* ('street') as last part of their name, which is usually a compound word or a phrase.

In a looser evaluation the categories were treated so that any correct marking of an entity regardless its boundaries was considered a hit. Four different location categories were joined to two: general location *<EnamexLocXxx>* and that of street names. End result was six different categories instead of eight. Table 2 shows evaluation results with loose evaluation. Recall and precision of the most frequent categories of person and location was now clearly higher, but still not very good.

| Label | P | R | F-score | Number of tags found |
|---|---|---|---|---|
| <EnamexPrsHum> | 63.30 | 53.69 | 58.10 | 2681 |
| <EnamexLocXxx> | 69.05 | 49.21 | 57.47 | 1541 |
| <EnamexLocStr> | 83.64 | 25.56 | 39.15 | 55 |
| <EnamexOrgEdu> | 51.72 | 47.62 | 49.59 | 58 |
| <EnamexOrgCrp> | 30.27 | 32.02 | 31.12 | 750 |
| <TimexTmeDat> | 73.85 | 12.62 | 21.56 | 673 |

**Table 2.** Evaluation results of FiNER with loose criteria and six categories

## 2.6     Results of ARPA

Our third evaluation was performed for a limited tag set with tools of the SeCo's ARPA. First only places were identified so that one location, *EnamexLocPpl*, was recognized. For this task, ARPA was first configured for the task of identifying place names in the data. As a first iteration, only the Finnish Place Name Registry[11] was used. After examining raw results from the test run, three issues were identified for further improvement. First, PNR contains only modern Finnish place names. To improve recall, three registries containing historical place names were added: 1) the Finnish spatiotemporal ontology SAPO (Hyvönen *et al.*, 2011) containing names of historic municipalities, 2) a repository of old Finnish maps and associated places from the 19th and early 20th Century, and 3) a name registry of places inside historic Karelia, which does not appear in PNR due to being ceded by Finland to the Soviet Union at the end of the Second World War (Ikkala *et al.*, 2016). To account for international place names, the names were also queried against the Geonames database[12] as well as Wikidata[13]. The contributions of each of these resources to the number of places identified in the final runs are shown in Table 3. Note that a single place name can be, and often was found in multiple of these sources.

| Source | Matches | Fuzzy matches |
|---|---|---|
| Karelian places | 461 | 951 |
| Old maps | 685 | 789 |
| Geonames | 1036 | 1265 |
| SAPO | 1467 | 1610 |
| Wikidata | 1877 | 2186 |
| PNR | 2232 | 2978 |

**Table 3.** Number of distinct place names identified using each source

Table 4 describes the results of location recognition with ARPA. With one exception (*New York*), only one word entities were discovered by the software.

| Label | P | R | F-score | Number of tags |
|---|---|---|---|---|
| <EnamexLocPpl/> | 39.02 | 53.24 | 45.03 | 2673 |
| </EnamexLocPpl> | 100.00 | 5.26 | 10.00 | 1 |
| <EnamexLocPpl> | 100.00 | 4.76 | 9.09 | 1 |

**Table 4.** Basic evaluation results for ARPA

---

A second improvement to the ARPA process arose from the observation that while recall in the first test run was high, precision was low. Analysis revealed this to be due to many names being both person names as well as places. Thus, a filtering step was added, that removed 1) hits identified as person names by the morphological analyzer and 2) hits that matched regular expressions catching common person name patterns found in the data (I. Lastname and FirstName LastName). However, sometimes this was too aggressive, ending up for example in filtering out also big cities like *Tampere* and *Helsinki*. Thus, in the final configuration, this filtering was made conditional on the size of the identified place, as stated in the structured data sources matched against.

Finally, as the amount of OCR errors in the target dataset was identified to be a major hurdle in accurate recognition, experiments were made with sacrificing precision in favor of recall through enabling various levels of Levenshtein distance matching against the place name registries. In this test, the fuzzy matching was done in the query phase after lexical processing. This was easy to do, but doing the fuzzy matching during lexical processing would probably be more optimal, as currently lemma guessing (which is needed because OCR errors are out of the lemmatizer's vocabulary) is extremely sensitive to OCR errors particularly in the suffix parts of words.

After the place recognition pipeline was finalized, a further test was done to test if the ARPA pipeline could be used for also person name recognition. The Virtual International Authority File was used as a lexicon of names, as it contains 33 million names for 20 million people. In the first run, the query simply matched all uppercase words against both first and last names in this database, while allowing for any number of initials to also precede such names matched. This way, the found names can't actually be always any more linked to strong identifiers, but for a pure NER task, recall is improved.

Table 5 shows results of this evaluation without fuzzy matching of names and Table 6 with fuzzy matching. Table 7 shows evaluation results with loose criteria without fuzzy matching and Table 8 loose evaluation with fuzzy matching.

| Label | P | R | F-score | Number of tags |
|---|---|---|---|---|
| <EnamexLocPpl/> | 58.90 | 55.59 | 57.20 | 1849 |
| </EnamexLocPpl> | 1.49 | 10.53 | 2.61 | 134 |
| <EnamexLocPpl> | 1.63 | 14.29 | 2.93 | 184 |
| <EnamexPrsHum/> | 30.42 | 27.03 | 28.63 | 2242 |
| </EnamexPersHum> | 83.08 | 47.39 | 60.35 | 656 |
| <EnamePersHum> | 85.23 | 43.80 | 57.87 | 738 |

**Table 5.** Evaluation results for ARPA: no fuzzy matching

| Label | P | R | F-score | Number of tags |
|---|---|---|---|---|
| <EnamexLocPpl/> | 47.38 | 61.82 | 53.64 | 2556 |
| </EnamexLocPpl> | 1.63 | 15.79 | 2.96 | 184 |
| <EnamexLocPpl> | 1.55 | 14.29 | 2.80 | 193 |
| <EnamexPrsHum/> | 9.86 | 66.79 | 17.18 | 3815 |
| </EnamexPersHum> | 63.07 | 62.97 | 63.01 | 1148 |
| <EnamePersHum> | 62.25 | 61.77 | 62.01 | 1425 |

**Table 6.** Evaluation results for ARPA: fuzzy matching

| Label | P | R | F-score | Number of tags |
|---|---|---|---|---|
| <EnamexPrsHum> | 63.61 | 45.27 | 52.90 | 3636 |
| <EnamexLocXxx> | 44.02 | 64.58 | 52.35 | 2933 |

**Table 7.** Evaluation results for ARPA with loose criteria: no fuzzy matching

| Label | P | R | F-score | Number of tags |
|---|---|---|---|---|
| <EnamexPrsHum> | 34.39 | 78.09 | 51.57 | 6388 |
| <EnamexLocXxx> | 44.02 | 64.58 | 52.35 | 2933 |

**Table 8.** Evaluation results for ARPA with loose criteria: fuzzy matching

Recall of recognition increases markedly in fuzzy matching, but precision deteriorates. More multipart location names are also recognized with fuzzy matching. In loose evaluation more tags are found but precision is not very good and thus the overall F-score is a bit lower than in the strict evaluation.

### 2.7 Results of other systems

Here we report briefly results of three other systems that we evaluated. These are Polyglot, a Finnish semantic tagger (Löfberg *et al.*, 2005) and Connexor's NER.

Polyglot[14] is a natural language pipeline that supports multilingual applications. Among Polyglot's tools is also NER. The NER models of Polyglot were trained on datasets extracted automatically from Wikipedia. Polyglot's NER supports currently 40 major languages.

Results of Polylot'sperformance in a loose evaluation with three categories are shown in table 9.

---

| Label | P | R | F-score | Number of tags found |
|---|---|---|---|---|
| <EnamexPrsHum> | 75.99 | 34.60 | 47.55 | 1433 |
| <EnamexLocXxx> | 83.56 | 32.28 | 46.57 | 821 |
| <EnamemOrgCrp> | 5.77 | 1.70 | 2.63 | 208 |

**Table 9.** Evaluation results of Polyglot with loose criteria and three categories

As can be seen from the figures, Polyglot has high precision with persons and locations, but quite bad recall, and F-scores are thus about 10 % units below FiNER's performance and clearly below performance of ARPA. With corporations Polyglot performs very poorly.

### 2.7.1 Results of a semantic tagger of Finnish

Semantic tagging can be briefly defined as a dictionary-based process of identifying and labeling the meaning of words in a given text according to some classification. The Finnish Semantic Tagger (FST) has its origins in Benedict, the EU-funded language technology project, the aim of which was to discover an optimal way of catering for the needs of dictionary users in modern electronic dictionaries by utilizing state-of-the-art language technology. FST is not a NER tool as such; it has first and foremost been developed for the analysis of full text.

The Finnish semantic tagger was developed using the English Semantic Tagger as a model. This semantic tagger was developed at the University Centre for Corpus Research on Language (UCREL) at Lancaster University as part of the UCREL Semantic Analysis System (USAS) framework (Rayson *et al.*, 2004), and both these equivalent semantic taggers were utilized in the Benedict project in the creation of a context-sensitive search tool for a new intelligent dictionary. In different evaluations the FST has been shown to be capable of dealing with most general domains which appear in a modern standard Finnish text. Furthermore, although the semantic lexical resources were originally developed for the analysis of general modern standard Finnish, evaluation results have shown that the lexical resources are also applicable to the analysis of both older Finnish text and the more informal type of writing found on the Web. In addition, the semantic lexical resources can be tailored for various domain-specific tasks thanks to the flexible USAS category system. The semantically categorized single word lexicon of the FST contains 46,225 entries and the multiword expression lexicon contains 4,422 entries (Piao *et al.*, 2016), representing all parts of speech. There are plans to expand the semantic lexical resources for the FST by adding different types of proper names in the near future in order to tailor them e.g. for NER tasks.

FST tags three different types of names: personal names, geographical names and other proper names. These are tagged with tags Z1, Z2, and Z3, respectively (Löfberg *et al.*, 2005). It does not distinguish first names and sure names, but it is able to tag first names of persons with male and female sub tags. As Z3 is a slightly vague cate-gory with names of organizations among others, we evaluate only categories Z1 and Z2, persons and locations.

FST tagged the list of 75,980 names as follows: it marked 5,569 names with tags Z1-Z3. Out of these 3,473 were tagged as persons, 2,010 as locations and rest as other names. It tagged 47,218 words with the tag Z99, which is a mark for lexically unknown words. Rest of the words, 23,193, were tagged with tags of common nouns. Thus FST's recall with the name list is not very high.

In table 10 we show results of FST's tagging of locations and persons in our evaluation data. As the tagger does not distinguish multipart names only loose evaluation was performed. We performed two evaluations: one with the words as they are, and the other with w→v substitution.

| Label | P | R | F-score | Number of tags found |
|---|---|---|---|---|
| \<EnamexPrsHum\> | 76.48 | 22.48 | 34.75 | 897 |
| \<EnamexLocXxx\> | 67.11 | 47.72 | 55.78 | 1420 |
| \<EnamexPrsHum\> w/v | 76.10 | 23.06 | 35.39 | 908 |
| \<EnamexLocXxx\> w/v | 69.66 | 51.34 | 59.12 | 1536 |

**Table 10.** Evaluation of FST tagger with loose criteria and two categories. W/v stands for w to v substitution in words.

Substitution of *w* with *v* decreased number of unknown words to FST with about 3 % units and has a noticeable effect on detection of locations and a small effect on persons. Overall locations are recognized better; their recognition with w/v substitution is slightly better than FiNER's and better than ARPA's overall. FST's recognition of persons is clearly inferior to that of FiNER and ARPA.

### 2.7.2 Results of Connexor's NER

Connexor Ltd. has provided different language technology tools, and among them is name recognition[15]. There is no documentation related to the software, but Connexor states on their Web pages that "using linguistic and heuristic methods, the names in the text can be tagged accurately". Software's name type repertoire is large; at least 31 different types of names are recognized. These are part of 9 larger categories like NAME.PER (persons), NAME.PRODUCT (products), NAME.GROUP (organizations), NAME.GPE (locations) etc. Boundaries of names are not tagged, so we perform only a loose evaluation.

As earlier, our interest is mainly in persons and locations. Connexor's tags NAME.GPE, NAME.GPE.City, NAME.GPE.Nation, NAME.GEO.Land and NAME.GEO.Water were all treated as \<EnamexLocXxx\>. NAME.PER, NAME.PER.LAW, NAME.PER.GPE, NAME.PER.Leader, NAME.PER.MED,

---

NAME.PER.TEO and NAME.PER.Title were all treated as <EnamexPrsHum>. All other tags were discarded. Results of Connexor's tagger are shown in Table 11.

| Label | P | R | F-score | Number of tags found |
|---|---|---|---|---|
| <EnamexPrsHum> | 44.86 | 76.02 | 56.40 | 5321 |
| <EnamexLocXxx> | 66.76 | 55.93 | 60.87 | 1802 |

**Table 11.** Evaluation of Connexor's tagger with loose criteria and two categories

Results show that Connexor's NE tagger is better with locations, but also persons are found well. Recall with persons is high, but low precision hurts overall performance. Data inspection shows that Connexor's tagger has a tendency to tag words beginning with upper case as persons. Locations are also mixed with persons many times.

### 2.8 Results overall

If we consider results of FiNER and ARPA overall, we can make the following observations. They both seem to find best two part person names, most of which consist of first name and last name. In strict evaluation ARPA appears better with locations than FiNER, but this is due to the fact that FiNER has a more fine-grained location tagging. With one location tag FiNER performs equally well as ARPA. In loose evaluation they both seem to find equally well locations and humans. FiNER finds educational organizations best, although they are scarce in the data. Corporations are also found relatively well, even though this category is prone to historical changes. FiNER is precise in finding two part street names, but recall in street name tagging is low. High precision is most probably due to common part *–katu* in street names: they are easy to recognize, if they are spelled right in the data. Low recall indicates bad OCR in street names.

One more caveat of FiNER's performance is in order. After we had achieved our evaluation results, we evaluated FiNER's context sensitivity with a small test. Table 12 shows effect of different contexts on FiNER's tagging for 320 names of municipalities. In the leftmost column are results, where only a name list was given to FiNER, in three other columns name of the municipality was changed from the beginning of a clause to middle and end. Results imply that there is context sensitivity in FiNER's tagging. With no context at all results are worst, and when the location is at the beginning of the sentence, FiNER misses also more tags than in other two positions. Overall it tags about two thirds of the municipality names as locations (LocXxx and LocGpl) in all the three context positions. High number of municipalities tagged as persons is partly understandable as names are ambiguous, but in many cases interpretation as a person is not well grounded. This phenomenon derives clearly from FiNER's tagging strategy that was explained at the end of section 2.1. At the beginning of the clause locations are not confused as much to persons, but this comes with a cost of more untagged names.

| No context, list of names | With context 1: location at the beginning | With context 2: location in the middle | With context 3: location at the end |
|---|---|---|---|
| 111 LocXxx | 151 LocXxx | 158 LocXxx | 159 LocXxx |
| 84 PrsHum | 66 PrsHum | 80 PersHum | 80 PersHum |
| 7 LocGpl | 56 LocGpl | 54 LocGpl | 54 LocGpl |
| 12 OrgCrp | 10 OrgCrp | 12 OrgCrp | 11 OrgCrp |
| 2 OrgTvr | 2 OrgTvr | 2 OrgTvr | 2 OrgTvr |
| 102 no tag | 35 no tag | 14 no tag | 14 no tag |

**Table 12.** FiNER's tagging for 320 names of municipalities with different positional context for the name

Same setting was tested further with 15,480 last names in three different clause positions. Positional effect with last name tagging was almost nonexistent, but amount of both untagged names and locative interpretations is high. 39 % of last names are tagged as PrsHum, 19.5 % are tagged as LocXxx, and about 34.6 % get no tag at all. The rest 7 % are in varying categories. Tagging of last names would probably be better, if first names were given together with last names. Isolated last names are more ambiguous.

Thus contextualization may have a minor effect on our results, as input text snippets were of different sizes, as mentioned in section 2.3. Person names may especially suffer, if first and last names are separated to different input snippets.

Out of our briefly evaluated tools FST was able to recognize locations slightly better than FiNER or ARPA in loose evaluation when w/v variation was neutralized. Connexor's tagger performed at the same level as FINER and ARPA in a loose evaluation. Its F-score with locations was the best performance overall. Polyglot performed clearly worst of all the systems.

## 3 Discussion

We have shown in this paper first evaluation results of NER for historical Finnish newspaper material from the 19[th] and early 20[th] century with two main tools, FiNER and SeCo's ARPA. Besides these two tools we evaluated shortly three other tools: a Finnish semantic tagger, Polyglot's NER and Connexor's NER. We were not able to train Stanford NER for Finnish. As far as we know, the tools we have evaluated constitute a comprehensive selection of tools that are capable of named entity recognition for Finnish, although not all of them are dedicated NER taggers.

Word level correctness of the whole digitized newspaper archive is approximately 70–75 % (Kettunen and Pääkkönen, 2016); the evaluation corpus had a word level correctness of about 73 %. Regarding this and the fact that FiNER and ARPA and other tools were developed for modern Finnish, the newspaper material makes a very difficult test for named entity recognition. It is obvious that the main obstacle of high class NER in this material is bad quality of the text. Also historical spelling variation has some effect, but it should not be that high, as late 19[th] century Finnish is not too far from modern Finnish and can be analyzed reasonably well with modern morphological tools (Kettunen and Pääkkönen, 2016). Morphological analyzers used in both FiNER and ARPA seem to be flexible and are able to analyze our low quality OCRed texts with a guessing mechanism, too. FST and Connexor's NER performed also quite well with morphology.

Evaluation results in this phase were not very good, best basic F-scores were ranging from 30 to 60 in the basic evaluation, and slightly better in a looser evaluation and with ARPA's fuzzy matching. To be able to estimate effect of bad OCR on the results, we made some unofficial extra trials with improved OCR material. We made tests with three versions of a 500,000 word text material that is different from our NER evaluation material but derives from the 19th century newspapers as well. One version was manually corrected OCR, another an old OCRed version and third a new OCRed version. Besides character level errors also word order errors have been corrected in the two new versions. For these texts we did not have a ground truth NE tagged version, and thus we could only count number of NER tags in different texts. With FiNER total number of tags increased from 23,918 to 26,674 (+11.5 % units) in the manually corrected text version. Number of tags increased to 26,424 tags (+10.5 % units) in the new OCRed text version. Most notable increase in the number of tags was in categories *EnamexLocStr* and *EnamexOrgEdu*. With ARPA results were even slightly better. ARPA recognized 10,853 places in the old OCR, 11,847 in the new OCR (+ 9.2 % units) and 13,080 (+20.5 % units) in the ground truth version of the text. Thus there is about a 10–20 % unit overall increase in the number of NER tags in both of the new better quality text versions in comparison to the old OCRed text with both taggers.

Another clear indication of effect of the OCR quality on the NER results is the following observation: when the words in all the correctly tagged FiNER *Enamex*es of the evaluation data are analyzed with Omorfi, only 14.3 % of them are unrecognized. With wrongly tagged FiNER *Enamex*es 26.3 % of the words are unrecognized by Omorfi. On tag wise level the difference is even clearer, as can be seen in recognition figures of Table 13 with words of locations and persons of FiNER, ARPA, FST and Connexor analyses (FiNER's analysis was reduced to a single location). Thus improvement in OCR quality will most probably bring forth a clear improvement in NER of the material.

|                                          | Locations | Persons |
|------------------------------------------|-----------|---------|
| FiNER right tag, word unrec. rate        | 6.3       | 12.8    |
| ARPA right tag, word unrec. rate         | 1.9       | 4.5     |
| FST right tag, word unrec. rate w/v      | 4.1       | 0.06    |
| Connexor right tag, word unrec.          | 10.22     | 25.01   |
| FiNER wrong tag, word unrec rate         | 38.3      | 34.0    |
| ARPA wrong tag, word unrec. rate         | 22.7      | 29.3    |
| FST wrong tag, word unrec. rate w/v      | 33.9      | 28.4    |
| Connexor wrong tag, word unrec. rate     | 53.45     | 57.39   |

**Table 13.** Word unrecognition percentages with rightly and wrongly tagged locations and persons – recognition with Omorfi 0.3

NER experiments with OCRed data in other languages show usually some improvement of NER when the quality of the OCRed data has been improved from very poor to somehow better (Packer *et al.*, 2010; Marrero *et al.*, 2013; Miller *et al.*, 2000). Results of Alex and Burns (2014) imply that with lower level OCR quality (below 70 % word level correctness) name recognition is harmed clearly. Packer *et al.* (2010) report partial correlation of Word Error Rate of the text and achieved NER result; their experiments imply that word order errors are more significant than character level errors. Miller *et al.* (2000) show that rate of achieved NER performance of a statistical trainable tagger degraded linearly as a function of word error rates. On the other hand, results of Rodriquez *et al.* (2012) show that manual correction of OCRed material that has 88–92 % word accuracy does not increase performance of four different NER tools significantly.

As the word accuracy of our material is low, it would be expectable, that somehow better recognition results would be achieved, if the word accuracy was round 80–90 % instead of 70–75 %. Our informal tests with different quality texts suggest this, too, as do the distinctly different unrecognition rates with rightly and wrongly tagged words.

Better quality for our texts may be achievable in the near future. Promising results in post correction of the Finnish historical newspaper data have been reported recently: two different correction algorithms developed in the FIN-CLARIN consortium achieved correction rate of 20–35 % (Silfverberg *et al.*, 2016). We are also progressing in re-OCRing tests of the newspaper data with open source OCR engine, Tesseract[16], and may be able to improve the OCR quality of our data (Kettunen *et al.*, 2016). Together improved OCR and post correction may yield 80+ % word level recognition for our data. Besides character level errors our material has also quite a lot of word order errors which may affect negatively the NER results (Packer *et al.*, 2010). Word order of the material may be improved in later processing of the XML ALTO and METS data, and this may also improve NER results. It would also be important that word splits due to hyphenation could be corrected in the data (Packer *et al.*, 2010)

---

[16] https://github.com/tesseract-ocr

Other suspected causes for poor NER performance could be due to 19[th] century Finnish spelling variation and perhaps also due to different writing conventions of the era. It is possible, for example, that the genre of 19[th] century newspaper writing differs from modern newspaper writing in some crucial aspects. Considering that both FiNER and ARPA are made for modern Finnish, our evaluation data is heavily out of their main scope (Poibeau and Kosseim, 2001), even if ARPA uses historical Finnish aware Omorfi and FiNER is able to guess unrecognized word forms.

One option for better NE recognition results is that we can use more historical language sensitive NER software. Such may become available, if the historically more sensitive version of morphological recognizer Omorfi can be merged with FiNER. Another possibility is to train a statistical name tagger described by Silfverberg (2015) with labeled historical newspaper material. Development work of a statistical NE tagger is underway in the FIN-CLARIN consortium. This version is targeted to domain of news of modern Finnish and is thus not directly applicable with our data, but as the tagger will be statistical, its domain may be changed with supervised learning. A new larger historical Finnish NER evaluation and teaching collection needs to be established for this purpose.

Finally, a note about usage of Named Entity Recognition is in order. Named Entity Recognition in itself is a tool that needs to be used to some useful purpose. In our case extraction of person and place names is primarily a tool for improving access to the Digi collection. After getting the recognition rate of the NER tool to an acceptable level, we need to decide, how we are going to use extracted names in Digi. Some exemplary suggestions are provided by archive of La Stampa and Trove Names (Mac Kim and Cassidy, 2015). La Stampa style usage of names provides informational filters after a basic search has been conducted. User can further look for persons, locations and organizations mentioned in the article results. This kind of approach enables browsing access to the collection and possibly also entity linking (Bates, 2007; Toms, 2000; McNamee *et al.*, 2011). Trove Names' name search takes the opposite approach: user searches first for names and then gets articles where the names occur. We believe that the La Stampa style of usage of names in the GUI of the newspaper collection is more informative and useful for users, as the Trove style can be achieved with the normal search function in the GUI of the newspaper collection.

If we consider possible uses of now evaluated NER tools in Digi, FiNER does so far only basic recognition and classification of names, which is the first stage (McNamee *et al.*, 2011). To be of general practical use names would need both intra document reference entity linking as well as multiple document reference entity linking (McNamee *et al.*, 2011; Ehrmann *et al.*, 2016). ARPA's semantic entity linking is of broader use, and entity linking has been used for example in the Europeana newspaper collection with names (Neudecker *et al.*, 2014; Hallo *et al.*, 2016). One more possible use for NER is usage with tagging and classification of images published in the newspapers. Most of the images (photos) have short title texts. It seems that many of the images represent locations and persons, with names of the objects mentioned in the image title. As image recognition and classifying of low quality print images may not be very feasible, image texts may offer a way to classify at least a reasonable part of the images. Along with NER also topic detection could be done to the image titles.

Our main emphasis with NER will be to use the names with the newspaper collection as a means to improve structuring, browsing and general informational usability of the collection. A good enough coverage of the names with NER needs to be achieved also for this use, of course. A reasonable balance of P/R should be found for this purpose, but also other capabilities of the software need to be considered. These remain to be seen later, if we are able to connect some type of functional NER to our historical newspaper collection's user interface.

## Acknowledgements

First and third author are funded by the EU Commission through its European Regional Development Fund, and the program Leverage from the EU 2014–2020.

Thanks to Heikki Kantola and Connexor Ltd. for providing the evaluation data of Connexor's NE tagger.

## References

Alex, B. and Burns, J. (2014), "Estimating and Rating the Quality of Optically Character Recognised Text", in *DATeCH '14 Proceedings of the First International Conference on Digital Access to Textual Cultural Heritage*, available at: http://dl.acm.org/citation.cfm?id=2595214 (accessed 10 October 2015).

Bates, M. (2007), "What is Browsing – really? A Model Drawing from Behavioural Science Research", *Information Research* 12, available at: http://www.informationr.net/ir/12-4/paper330.html (accessed 1 June 2016).

Bremer-Laamanen, M-L. (2014), "In the Spotlight for Crowdsourcing", *Scandinavian Librarian Quarterly*, Vol. 46, No. 1, pp. 18–21.

Crane, G. and Jones, A. (2006), "The Challenge of Virginia Banks: An Evaluation of Named Entity Analysis in a 19th-Century Newspaper Collection", in *Proceedings of JCDL'06, June 11–15, 2006, Chapel Hill, North Carolina, USA*, available at: http://citeseerx.ist.psu.edu/viewdoc/download?doi=10.1.1.91.6257&rep=rep1&type=pdf (accessed 1 June 2016).

Ehrmann, M., Nouvel, D. and Rosset, S. (2016), "Named Entity Resources – Overview and Outlook", in *LREC 2016, Tenth International Conference on Language Resources and Evaluation*, available at http://www.lrec-conf.org/proceedings/lrec2016/pdf/987_Paper.pdf (accessed June 15 2016).

Hallo, M., Luján-Mora, S., Maté, A. and Trujillo, J. (2016), "Current State of Linked Data in Digital Libraries", *Journal of Information Science*, Vol. 42, No. 2, pp. 117–127.

Hyvönen, E., Tuominen, J., Kauppinen T. and Väätäinen, J. (2011), "Representing and Utilizing Changing Historical Places as an Ontology Time Series", in Ashish, N. and Sheth, V. (Eds.) *Geospatial Semantics and Semantic Web: Foundations, Algorithms, and Applications*, Springer US, pp. 1–25.

Ikkala, E., Tuominen, J. and Hyvönen, E. (2016), "Contextualizing Historical Places in a Gazetteer by Using Historical Maps and Linked Data", *in Digital Humanities 2016: Conference Abstracts*, Jagiellonian University & Pedagogical University, Kraków, pp. 573-577, available at: http://dh2016.adho.org/abstracts/39 (accessed October 1 2016).

Järvelin, A., Keskustalo, H., Sormunen, E., Saastamoinen, M. and Kettunen, K. (2015), "Information Retrieval from Historical Newspaper Collections in Highly Inflectional Languages: A Query Expansion Approach", *Journal of the Association for Information Science and Technology,* available at: http://onlinelibrary.wiley.com/doi/10.1002/asi.23379/epdf (accessed January 12 2016).

Kettunen, K., Honkela, T., Lindén, K., Kauppinen, P., Pääkkönen, T. and Kervinen, J. (2014), "Analyzing and Improving the Quality of a Historical News Collection using Language Technology and Statistical Machine Learning Methods", in *Proceedings of IFLA 2014, Lyon (2014)*, available at: http://www.ifla.org/files/assets/newspapers/Geneva_2014/s6-honkela-en.pdf (accessed March 15 2015).

Kettunen, K. and Pääkkönen, T. (2016), "Measuring Lexical Quality of a Historical Finnish Newspaper Collection – Analysis of Garbled OCR Data with Basic Language Technology Tools and Means", in *LREC 2016, Tenth International Conference on Language Resources and Evaluation*, available at http://www.lrec-conf.org/proceedings/lrec2016/pdf/17_Paper.pdf (accessed 15 June 2016).

Kettunen, K., Pääkkönen, T. and Koistinen, M. (2016), "Between Diachrony and Synchrony: Evaluation of Lexical Quality of a Digitized Historical Finnish Newspaper and Journal Collection with Morphological Analyzers", in: Skadiņa, I. and Rozis, R. (Eds.), *Human Language Technologies – The Baltic Perspective*, IOS Press, pp. 122–129. Available at: http://ebooks.iospress.nl/volumearticle/45525 (accessed October 12 2016).

Kokkinakis, D., Niemi, J., Hardwick, S., Lindén, K., and Borin. L. (2014), "HFST-SweNER – a New NER Resource for Swedish". in: *Proceedings of LREC 2014*, available at: http://www.lrec-conf.org/proceedings/lrec2014/pdf/391_Paper.pdf (accessed 15 June 2016).

Löfberg, L., Piao, S., Rayson, P., Juntunen, J-P, Nykänen, A. and Varantola, K. (2005), "A semantic tagger for the Finnish language", available at http://eprints.lancs.ac.uk/12685/1/cl2005_fst.pdf (accessed 15 June 2016).

Lindén, K., Axelson, E., Drobac, S., Hardwick, S., Kuokkala, J., Niemi, J., Pirinen, T.A. and Silfverberg, M. (2013) "HFST—a System for Creating NLP Tools", in Mahlow, C., Piotrowski, M. (eds.) *Systems and Frameworks for Computational Morphology. Third International Workshop, SFCM 2013*, Berlin, Germany, September 6, 2013 Proceedings, pp. 53–71.

Lopresti, D. (2009), "Optical character recognition errors and their effects on natural language processing", *International Journal on Document Analysis and Recognition,* Vol. 12**,** No. 3, pp. 141–151.

Mac Kim, S. and Cassidy, S. (2015), "Finding Names in Trove: Named Entity Recognition for Australian", in *Proceedings of Australasian Language Technology Association Workshop*, available at: https://aclweb.org/anthology/U/U15/U15-1007.pdf (accessed August 10 2016).

Mäkelä, E. (2014), "Combining a REST Lexical Analysis Web Service with SPARQL for Mashup Semantic Annotation from Text", In Presutti, V. et al. (Eds.), *The Semantic Web: ESWC 2014 Satellite Events*. Lecture Notes in Computer Science, vol. 8798, Springer, pp. 424–428.

Manning, C. D., Schütze, H. (1999) *Foundations of Statistical Language Processing*. The MIT Press, Cambridge, Massachusetts.

Marrero, M., Urbano, J., Sánchez-Cuadrado, S., Morato, J. and Gómez-Berbís, J.M. (2013), "Named Entity Recognition: Fallacies, challenges and opportunities", *Computer Standards & Interfaces* Vol. 35 No. 5, pp.  482–489.

McNamee, P., Mayfield, J.C., and Piatko, C.D. (2011), "Processing Named Entities in Text", *Johns Hopkins APL Technical Digest*, Vol. 30 No. 1, pp. 31–40.

Miller, D., Boisen, S., Schwartz, R. Stone, R. and Weischedel, R. (2000), "Named entity extraction from noisy input: Speech and OCR", in *Proceedings of the 6th Applied Natural Language Processing Conference*, 316–324, Seattle, WA, available at: http://www.anthology.aclweb.org/A/A00/A00-1044.pdf (accessed 10 August 2016).

Nadeau, D., and Sekine, S. (2007), "A Survey of Named Entity Recognition and Classification", *Linguisticae Investigationes,* Vol. 30 No. 1, pp. 3–26.

Neudecker, C., Wilms, L., Faber, W. J., and van Veen, T. (2014), "Large-scale Refinement of Digital Historic Newspapers with Named Entity Recognition", In *Proceedings of IFLA 2014,* available at: http://www.ifla.org/files/assets/newspapers/Geneva_2014/s6-neudecker_faber_wilms-en.pdf (accessed June 10 2016).

Neudecker, C. (2016), "An Open Corpus for Named Entity Recognition in Historic Newspapers", in *LREC 2016, Tenth International Conference on Language Re-* *sources and Evaluation*, available at http://www.lrec-conf.org/proceedings/lrec2016/pdf/110_Paper.pdf (accessed June 17 2016).

Pääkkönen, T., Kervinen, J., Nivala, A., Kettunen, K. and Mäkelä E. (2016), "Exporting Finnish Digitized Historical Newspaper Contents for Offline Use", *D-Lib Magazine*, July/August, available at http://www.dlib.org/dlib/july16/paakkonen/07paakkonen.html (accessed August 15 2016).

Packer, T., Lutes, J., Stewart, A., Embley, D., Ringger, E., Seppi, K. and Jensen, L. S. (2010), "Extracting Person Names from Diverse and Noisy OCR Text", in *Proceedings of the fourth workshop on Analytics for noisy unstructured text data. Toronto, ON, Canada*: ACM, available at: http://dl.acm.org/citation.cfm?id=1871845 (accessed May 10 2016).

Poibeau, T. and Kosseim, L. (2001), "Proper Name Extraction from Non-Journalistic Texts", *Language and Computers,* Vol. 37 No. 1, pp. 144–157.

Rayson, P., Archer, D., Piao, S. L. and McEnery, T. (2004), "The UCREL semantic analysis system" in *Proceedings of the workshop on Beyond Named Entity Recognition Semantic labelling for NLP tasks in association with 4th International Conference on Language Resources and Evaluation (LREC 2004), 25th May 2004, Lisbon, Portugal,* pp. 7-12. Available at: http://www.lancaster.ac.uk/staff/rayson/publications/usas_lrec04ws.pdf (accessed August 10 2016).

Rodrigues, K.J., Bryant, M., Blanke, T. and Luszczynska, M. (2012), "Comparison of Named Entity Recognition Tools for raw OCR text", in: *Proceedings of KONVENS 2012 (LThist 2012 wordshop), Vienna* September 21, pp. 410–414.

Piao, S., Rayson, P., Archer, D., Bianchi, F., Dayrell, C., El-Haj, M., Jiménez, R.-M., Knight, D., Kren, M., Löfberg, L., Nawab, R.M.A., Shafi, J., The, P.L. and Mudraya, O. (2016), "Lexical Coverage Evaluation of Large-scale Multilingual Semantic Lexicons for Twelve Languages", in *LREC 2016, Tenth International Conference on Language Resources and Evaluation*, available at: http://www.lrec-conf.org/proceedings/lrec2016/pdf/257_Paper.pdf (accessed August 10 2016).

Silfverberg, M., Kauppinen, P., and Linden, K. (2016), "Data-Driven Spelling Correction Using Weighted Finite-State Methods", in *Proceedings of the ACL Workshop on Statistical NLP and Weighted Automata,* pp. 51–59, available at: https://aclweb.org/anthology/W/W16/W16-2406.pdf (accessed August 20 2016).

Silfverberg, M. (2015), "Reverse Engineering a Rule-Based Finnish Named Entity Recognizer", paper presented at Named Entity Recognition in Digital Humanities Workshop, June 15, Helsinki available at: https://kitwiki.csc.fi/twiki/pub/FinCLARIN/KielipankkiEventNERWorkshop2015/Silfverberg_presentation.pdf (accessed April 5 2016).

Tkachenko, A., Petmanson, T., and Laur, S. (2013), "Named Entity Recognition in Estonian", in *Proceedings of the 4th Biennial International Workshop on Balto-Slavic Natural Language Processing*, pp. 78–83, available at: http://aclweb.org/anthology/W13-24 (accessed May 10 2016).

Toms, E.G. (2000), "Understanding and Facilitating the Browsing of Electronic Text", *International Journal of Human-Computer Studies*, Vol. 52 No. 3, pp. 423–452.